\documentclass[]{interact}

\usepackage{epstopdf}% To incorporate .eps illustrations using PDFLaTeX.
\usepackage[caption=false]{subfig}% Support for small, `sub' figures and tables

\theoremstyle{plain}% Theorem-like structures provided by amsthm.sty

\theoremstyle{definition}

\theoremstyle{remark}

\usepackage[numbers,sort&compress]{natbib}
\bibpunct[, ]{[}{]}{,}{n}{,}{,}
\renewcommand\bibfont{\fontsize{10}{12}\selectfont}
\makeatletter
\def\NAT@def@citea{\def\@citea{\NAT@separator}}
\makeatother

\usepackage{algorithm,algorithmic}
\newcommand{\eg}{\textit{e.g.,}~}
\newcommand{\ie}{\textit{i.e.,}~}

\newcommand*{\sref}[1]{\S\ref{#1}}            % section
\newcommand*{\apref}[1]{Appendix~\ref{#1}}            % section
\newcommand*{\aref}[1]{Algorithm~\ref{#1}}  % algorithm
\newcommand*{\tref}[1]{\text{TABLE~\ref{#1}}}   % table
\newcommand*{\fref}[1]{\text{Fig.~\ref{#1}}}
\newcommand*{\eref}[1]{\text{Eq.(\ref{#1})}}     %.

\usepackage[inline]{enumitem}\setlist[enumerate]*{label=(\roman*)}

\usepackage{url}
\usepackage{siunitx}

\usepackage[table]{xcolor}
\newcolumntype{C}[1]{>{\centering\arraybackslash}p{#1}}

\begin{document}

% \articletype{ARTICLE TEMPLATE}% Specify the article type or omit as appropriate

\title{
Self-Augmented Robot Trajectory: Efficient Imitation Learning via Safe Self-augmentation with Demonstrator-annotated Precision
}

\author{
\name{
Hanbit~Oh\textsuperscript{a*$\dagger$},
\thanks{* Equal contribution.}
\thanks{$\dagger$ Corresponding author.}
\thanks{CONTACT: oh.hanbit.oe9@aist.go.jp}
Masaki~Murooka\textsuperscript{b*},
Tomohiro~Motoda\textsuperscript{a},
Ryoichi~Nakajo\textsuperscript{a} and
Yukiyasu~Domae\textsuperscript{a}
} 
\affil{\textsuperscript{a} Embodied AI Research Team (EART), Artificial Intelligence Research Center, National Institute of Advanced Industrial Science and Technology (AIST), Japan; \textsuperscript{b}CNRS-AIST JRL (Joint Robotics Laboratory), IRL, National Institute of Advanced Industrial Science and Technology (AIST), Japan}}

\maketitle

\begin{abstract}
Imitation learning is a promising paradigm for training robot agents; however, standard approaches typically require substantial data acquisition---via numerous demonstrations or random exploration---to ensure reliable performance. Although exploration reduces human effort, it lacks safety guarantees and often results in frequent collisions---particularly in clearance-limited tasks (\eg peg-in-hole)---thereby, necessitating manual environmental resets and imposing additional human burden. This study proposes \textit{Self-Augmented Robot Trajectory (SART)}, a framework that enables policy learning from a single human demonstration, while safely expanding the dataset through autonomous augmentation. SART consists of two stages: (1) \textit{human teaching only once}, where a single demonstration is provided and precision boundaries---represented as spheres around key waypoints---are annotated, followed by one environment reset; (2) \textit{robot self-augmentation}, where the robot generates diverse, collision-free trajectories within these boundaries and reconnects to the original demonstration. This design improves the data collection efficiency by minimizing human effort while ensuring safety. Extensive evaluations in simulation and real-world manipulation tasks show that SART achieves substantially higher success rates than policies trained solely on human-collected demonstrations. 
Video results available at \url{https://sites.google.com/view/sart-il}.
\end{abstract}

\begin{keywords}
Imitation Learning; Learning from Human Demonstration; Data Augmentation
\end{keywords}

%%%%%%%%%%%%%%%%%%%%%%%%%%%%%%%%%%%%%%%%%%%%%%%%%%%%%%%%%%%%%%%%%%%%%%%%%%%%%%%%
\section{Introduction}
Imitation learning (IL) has emerged as a compelling technique for robot automation that allows agents to acquire skills from human demonstrations rather than explicit programming \cite{osa2018algorithmic}. This approach has demonstrated remarkable success across a range of robotic tasks and does not require a predefined model of the environment \cite{o2024rtx, kim2025openvla}.

\begin{figure}[t]
    \centering
    \includegraphics[width=\textwidth]{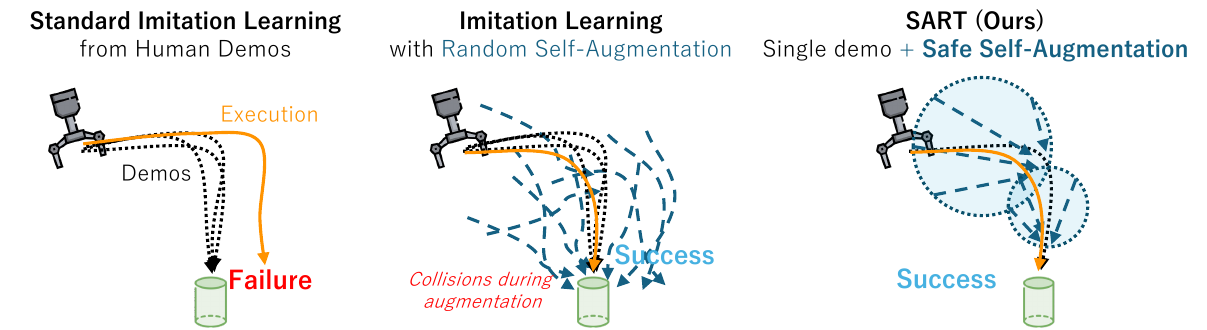}
    \caption{Illustrative example comparing standard imitation learning (left), imitation learning with random self-augmentation (center), and the proposed self-augmented robot trajectory (SART) method (right). Standard imitation learning suffers from covariate shift due to limited variation in demonstrations, often leading to failures in grasping tasks; moreover, failures typically require frequent environment resets. Random self-augmentation---such as random exploration via reinforcement learning---can expand data coverage. However, collisions during exploration create unsafe behaviors and force repeated environment resets. In contrast, SART generates diverse, collision-free trajectories from a single human demonstration within annotated precision boundaries, eliminating the need for resets while reducing covariate shift and enabling successful policy execution.}
    \label{fig:idea}
\end{figure}

Despite these advantages, many existing IL methods depend on extensive datasets of human demonstrations, the collection of which is both labor intensive and time consuming \cite{o2024rtx, kim2025openvla}. To mitigate this bottleneck, recent efforts have explored robotic data augmentation, that is, automatically expanding the training set by allowing robots to explore and generate supplementary data \cite{ho2016gail,papagiannis2022sil,ankile2024juicer,papagiannis2025miles}. Although this strategy can effectively reduce the need for human involvement, methods based on random exploration are ill-suited to clearance-limited tasks, where uninformed movements often lead to collisions and necessitate manual environment resets, as illustrated in \fref{fig:idea}. As a result, the overall human effort may not be significantly reduced. Addressing these issues often requires additional assumptions regarding the environment or task structure, which limit the general applicability of IL. Addressing these issues often requires additional environmental or structural assumptions, limiting the generality of IL. 
For example, Papagiannis et al.~\cite{papagiannis2025miles} reduce collisions by constraining exploration to a narrow, fixed region around each waypoint—effective for ensuring safety near contact-critical areas but overly restrictive in open spaces, leading to inefficient exploration. This motivates the development of a method that achieves a better balance between safety and efficiency across varying task conditions.

To this end, the self-augmented robot trajectory (SART) is proposed, which is a safe and efficient imitation learning framework that enables policy learning from a single human demonstration augmented by automatically generated data within explicitly annotated precision boundaries, as illustrated in \fref{fig:idea}. In SART, a human expert provides a single demonstration trajectory for the task and annotates safety regions—represented as spheres—at each key waypoint. These regions define the permissible operating space for the robot's end effector, allowing autonomous augmentation, which is both safe and effective. 
Extensive evaluations in both simulation and real-world manipulation tasks demonstrate that SART achieves substantially higher success rates than baselines, while also reducing the overall human duration required for dataset collection.

In summary, the primary contributions are as follows:
\begin{itemize}
    \item Self-augmented robot trajectory (SART) is introduced, which is a framework that enables robots to safely augment their own training data within human-annotated spatial boundaries derived from a single demonstration.
    \item By autonomously generating diverse trajectories within these safe boundaries, SART efficiently expands the demonstration dataset and yields more robust policies than those trained on human demonstrations alone.
    \item Extensive evaluations in both simulation and physical robot experiments show that SART improves learning efficiency and achieves higher success rates than fully human-supervised baselines in clearance-limited tasks (\eg peg-in-hole, door opening, and lid opening).
\end{itemize}

\section{Related Works}
\subsection{IL from Human Demonstration}
IL is an attractive paradigm for automating robotic manipulation tasks. The predominant approach within this domain is behavioral cloning (BC), in which a human expert provides example task executions as state-action trajectories \cite{bojarski2016end, osa2018algorithmic}. These data were used to train a control policy via supervised learning to mimic the demonstrated behavior. However, a well-known limitation of the BC is its susceptibility to distributional mismatch, which often encounters novel states that are not covered in demonstrations, leading to compounding errors and poor generalization \cite{ross2011dagger}.

To mitigate this, some methods have aggregated large-scale demonstration datasets to broaden the policy coverage and robustness \cite{o2024rtx, kim2025openvla}. However, the extensive human effort required for data collection places a significant burden on practitioners, making this approach costly and time-consuming. Alternatively, interactive data collection strategies \cite{celemin2022interactive} aim to reduce distributional mismatches, for example, by introducing perturbations during demonstrations to broaden the demonstration distribution \cite{laskey2017dart, oh2023bdi, tahara2022disturbance, tahara2023disturbance} or by deploying poor robot policies under human supervision with real-time corrective feedback \cite{ross2011dagger, Oh2024DPIIL, Ohchoa2025ISPIL}. While these approaches can effectively reduce distributional mismatch and improve policy robustness, they still demand ongoing human involvement, cognitive attention, and tolerance for disturbances, highlighting the need for methods that can minimize human workload.

\subsection{Efficient IL via Robotic Data Augmentation}
To reduce human effort while improving scalability, recent research has explored robotic data augmentation, where robots autonomously expand and diversify training datasets. A complementary approach is reinforcement learning (RL)-based imitation, which learns from one or a few demonstrations by optimizing the trajectory-level similarity to the reference behavior, including adversarial occupancy matching \cite{ho2016gail} and optimal transport with Sinkhorn objectives \cite{papagiannis2022sil}. Despite the strong results, these methods typically require iterative on-policy exploration to evaluate and update the policy, which in physical manipulation often necessitates repeated environment resets and substantial operator supervision and can further induce unintended contacts in clearance-limited settings during data collection.

Simulation-based approaches leverage the flexibility of virtual environments to generate numerous synthetic demonstrations from a limited set of seed trajectories \cite{mandlekar2023mimicgen, garrett2024skillmimicgen, jiang2024dexmimicgen, ankile2024juicer}. A representative example is the \textit{MimicGen-family} method ~\cite{mandlekar2023mimicgen, garrett2024skillmimicgen, jiang2024dexmimicgen}, which decomposes a few human demonstrations into object-centric subtasks and retargets them in new contexts through spatial and kinematic transformations. Similarly, JUICER increases the dataset density by synthesizing correction data from reversed disassembly actions \cite{ankile2024juicer}. These approaches produced successful trajectories across novel configurations and various robot morphologies, making them highly effective for simulations. Although MimicGen has also been demonstrated in real-world scenarios~\cite{mandlekar2023mimicgen}, these methods are fundamentally simulation-first and face practical obstacles in physical settings, each of which still requires manual randomization of the object placement, and reliable real-time object pose estimation is indispensable. Consequently, their applicability to real-world learning remains limited compared with their efficiency in simulations.

Although simulation-based augmentation can be effective, it often encounters scalability and safety constraints in actual environments. Alternatively, recent studies have used vision-based generative techniques to synthesize corrective demonstrations. Using models like neural radiance fields and diffusion models to generate plausible, novel observations and actions, these approaches enrich data diversity for robust policy learning \cite{zhou2023nerf, zhang2024diffusion}. Nonetheless, the utility of such synthesized data depends strongly on the fidelity of the underlying generative models; limited geometric accuracy and the absence of explicit kinematic/physical constraints can yield samples that are visually plausible, yet physically or kinematically infeasible for execution.

By contrast, self-supervised physical-world augmentation densifies the demonstration space through autonomous exploration by robots. Approaches like MILES~\cite{papagiannis2025miles} perturb a single demonstration trajectory and learn recovery behaviors from out-of-demonstration states, thereby enabling efficient augmentation without additional human demonstrations. However, MILES uniformly defines fixed-size augmentation regions around all waypoints without accounting for state-dependent precision. In low-precision regions (\eg open workspace areas), this often results in insufficient augmented coverage, exacerbating the distribution mismatch during policy execution.

The proposed method, SART, directly addresses these safety limitations by leveraging human-annotated collision-free boundaries. By restricting augmentation volume in high-precision regions, SART ensures safe exploration near contact-sensitive states. Conversely, in low-precision regions, such as open workspace areas, the boundaries can be expanded to fully exploit the available free space. This allows the robot to collect more diverse trajectories in states where variability is acceptable, thereby mitigating potential distribution mismatches at the test time and improving robustness.

\section{Preliminaries: IL from Human Demonstration} 
\label{sec:preliminaries}

An IL setting in which a robot learns manipulation tasks from human demonstrations is considered. Following previous work~\cite{mandlekar2023mimicgen, papagiannis2025miles}, the robot was equipped with a wrist-mounted RGB-D camera rigidly attached to its end-effector (EE), providing a consistent egocentric view. It was assumed that only the task-relevant object was visible during the demonstrations, allowing observations and actions to be defined relative to that object. This hand–eye configuration eliminates the need for external cameras and calibration, simplifying the setup while ensuring that perception remains aligned with the robot’s actions.

Demonstration $\mathcal{D}$ consists of a sequence of $T$ waypoints: $\mathcal{D}=\{\mathcal{W}_t\}_{t=1}^{T}$, where each waypoint $\mathcal{W}_t$ at time $t$ is composed of an observation $\mathbf{o}_t$ and the associated expert action $\mathbf{a}^*_t$. 
An observation $\mathbf{o}_t$ includes an RGB image from the wrist-mounted camera and gripper state (open or closed). Action $\mathbf{a}_t$ contains the desired 6-DoF pose and gripper state at the next time step, expressed relative to the EE’s pose at $t - 1$. The gripper state may be omitted when the task does not necessarily involve grasping (\eg door-opening task).

IL aims to learn a parametric policy $\pi_{\theta}(\mathbf{a}_t \mid \mathbf{o}_t)$ that performs actions $\mathbf{a}_t$ for a given observation $\mathbf{o}_t$ by imitating the expert policy $\pi_{\theta^*}(\mathbf{a}_t^* \mid \mathbf{o}_t)$. To formalize this objective, a surrogate loss function $J$ is defined to measure the action discrepancy between the learning and expert policies over a demonstration $\mathcal{D}$:
\begin{equation}
J(\pi_{\theta}, \pi_{\theta^*} \mid \mathcal{D}) 
= \sum_{t=1}^{T} \mathbb{E}_{\mathbf{a}_t^*\sim\pi_{\theta^*}} 
\left[ \left\| \pi_{\theta}(\mathbf{o}_t) - \mathbf{a}_t^* \right\|_2^2 \right].
\end{equation}
The parameter $\theta^L$ of the learned policy is then optimized by minimizing the expected surrogate loss over the demonstration data $\mathcal{D}$ sampled by a human expert:
\begin{equation}
\theta^L = \arg\min_{\theta} 
\mathbb{E}_{\mathcal{D}\sim \pi_{\theta^*}} 
\left[ J(\pi_{\theta}, \pi_{\theta^*} \mid \mathcal{D}) \right].
\label{eq:bc_objective}
\end{equation}

While standard ILs, such as BC \cite{bojarski2016end}, are simple and effective, they have an intrinsic limitation--- they rely solely on human demonstrations, resulting in limited observation coverage and a potential covariate shift \cite{ross2011dagger}. Moreover, collecting a sufficiently diverse set of demonstrations to improve the generalizability requires considerable effort \cite{o2024rtx}.

The following section introduces a method that addresses this limitation by reducing human effort, requiring only a single demonstration and a set of annotated precision boundaries that can be provided with minimal effort. The robot then autonomously augments its training data within these human-defined boundaries, enabling an efficient and safe expansion of the demonstration.

\section{Proposed Method: Self-Augmented Robot Trajectory (SART)}
\begin{figure}[t]
    \centering
    \includegraphics[width=\linewidth]{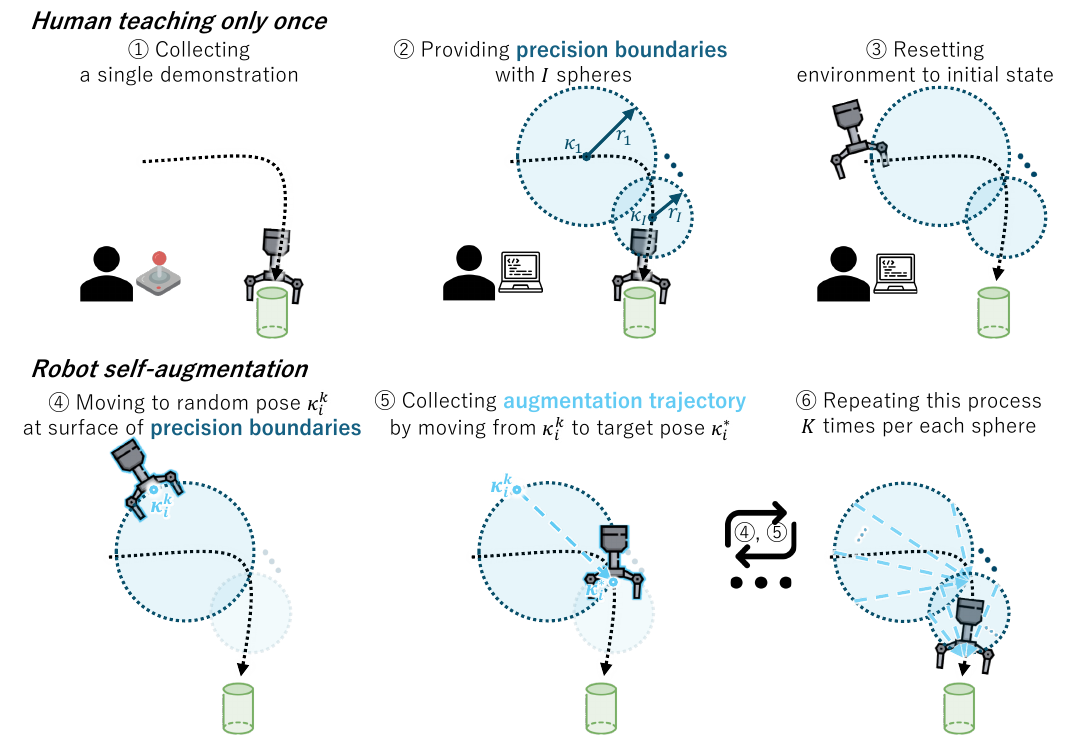}
    \caption{Overview of SART: A human provides a single demonstration and explicitly annotates precision boundaries around key waypoints. The robot then autonomously augments its dataset by exploring within these boundaries, generating diverse, collision-free trajectories used to learn a robust policy.}
    \label{fig:overview}
\end{figure}
% Brief overview
This section introduces the proposed framework, SART, which enables safe and efficient policy learning from a single human demonstration. This method comprises the following two critical components: 
\begin{enumerate*}
    \item A human provides a single demonstration along with annotated precision boundaries around key waypoints;
    \item The robot autonomously augments its training data by exploring within these boundaries to generate diverse, collision-free trajectories for policy learning.
\end{enumerate*}

\subsection{Human Teaching Only Once}

The first phase of SART involves a brief interaction with a human expert. The humans provided a single demonstration by teleoperating the robot to execute the task. After recording the demonstration, the environment is reset to its initial configuration to ensure consistency in the augmentation. This is the only reset required throughout SART because all subsequent data collection is performed autonomously by the robot.

Subsequently, a human annotates the precision boundaries along the demonstration trajectory. An interactive annotation interface is developed to assist with this process. This interface creates a virtual 3D point-cloud environment using RGB-D data captured by a wrist-mounted camera during the demonstration. It visualizes a virtual robot using a robot geometric model (\eg an URDF model) and displays the XYZ 3D end-effector pose of each waypoint, enabling humans to intuitively inspect the environment and assess the spatial precision required at each point. See Appendix~\sref{app:annotation} for details.

\textit{Precision boundaries} are represented as a set of $I$ spheres (one per selected key waypoint). This spherical shape was selected because of its simplicity, task independence, and computational efficiency, which are widely exploited for fast collision checking in motion planning \cite{zucker2013chomp}. A human selects $I$ key waypoints from the demonstration. Let the $i$-th key pose be $\kappa_{i} = \{\mathbf{c}^{\mathrm{pos}}_i, \mathbf{C}^{\mathrm{ori}}_i\} \in \mathrm{SE}(3)$, where $\mathbf{c}^{\mathrm{pos}}_i$ and $\mathbf{C}^{\mathrm{ori}}_i$ denote the end-effector’s position and orientation, respectively. A spherical boundary of radius $r_i$ is centered at $\mathbf{c}^{\mathrm{pos}}_i$, where the sphere’s surface defines the allowable variation for safe augmentation. In practice, the interface steps through a waypoint sequence with simple keyboard inputs, graphically rendering each sphere $(\kappa_i, r_i)$ such that the annotator can intuitively adjust $r_i$ while verifying the collision avoidance.

\subsection{Robot Self-Augmentation}

With the annotated boundaries defined, the robot performs self-augmentation by sampling random 3D poses on the surface of each annotated precision sphere and generating motions that are reconnected to the original demonstration at the \textit{convergence waypoint}. 
A convergence waypoint is defined as the first waypoint on the demonstration trajectory that lies outside the corresponding precision sphere. 
The resulting trajectories were recorded and repeated $K$ times per sphere to generate diverse data that converged back to the original demonstration path. This procedure is described in detail below.

To sample a random 3D position $\mathbf{p}_{i}^{k} \in \mathbb{R}^3$ on the surface of the sphere, the robot uses spherical coordinates by drawing two angles— $\psi \sim \mathcal{U}[0, 2\pi]$ and $\phi \sim \mathcal{U}[0, \pi]$, where $\psi$ is the azimuthal angle, and $\phi$ is the polar angle. The sampled point $\mathbf{p}_{i}^{k}$ is then computed as
\begin{align}
\mathbf{p}_{i}^{k} = \mathbf{c}^{\mathrm{pos}}_i + r_i
\begin{bmatrix}
\sin(\phi)\cos(\psi) \\
\sin(\phi)\sin(\psi) \\
\cos(\phi)
\end{bmatrix}.
\label{eq:rand_pos}
\end{align}
To augment 3D orientation, a random rotational perturbation was applied to the original demonstration trajectory. The maximum angular deviation is defined as
\begin{align}
\alpha_{i}^{\mathrm{max}} = \nu_r r_i,
\end{align}
where $\nu_r$ denotes a constant scale factor.
A random rotation angle $\alpha_i \sim \mathcal{U}[-\alpha_i^{\mathrm{max}}, \alpha_i^{\mathrm{max}}]$ was sampled, and a random axis was uniformly sampled on the unit sphere. A rotation matrix $\Delta \mathbf{R}_{i}^{k} \in \mathbb{R}^{3 \times 3}$ was then constructed from these angles and axes. The sampled rotation is given by
\begin{align}
\mathbf{R}_{i}^{k} = \mathbf{C}^{\mathrm{ori}}_i \, \Delta \mathbf{R}_{i}^{k}.
\label{eq:rand_ori}
\end{align}

The convergence waypoint was defined as the first waypoint outside the annotated sphere. Let $t'$ denote the time step of a selected waypoint. The convergence waypoint $\kappa_{i}^{*} =\{\mathbf{c}^{\mathrm{pos}}_{t^*}, \mathbf{C}^{\mathrm{ori}}_{t^*}\} \in \mathcal{W}_{t^*}$ is subsequently computed as the earliest waypoint on the demonstration trajectory that satisfies:
\begin{align}
t^* = \min \left\{\, t \;\middle|\; t' < t \leq T,\; \left\| \mathbf{c}^{\mathrm{pos}}_{t} - \mathbf{c}^{\mathrm{pos}}_{i} \right\|_2^2 > r_i^2 \right\}.
\label{eq:target_pose}
\end{align}

After computing the convergence waypoint $\kappa_{i}^{*}$, the robot executes a straight-line motion from $\kappa_{i}^{*}$ to the sampled pose $\kappa_{i}^{k}=(\mathbf{p}_{i}^{k}, \mathbf{R}_{i}^{k})$ over a fixed duration $\delta$. The recorded sequence is then reversed and concatenated with the demonstration. Starting from the convergence waypoint ensures precise alignment with the demonstration at the connection point and allows the joint configuration to match that of the original trajectory, preserving both the end-effector and joint continuity. 
Accordingly, each augmented trajectory is constructed by concatenating two parts: the automatically generated segment $\tau^{\mathrm{aug}}_{i,k}$ and the subsequent portion of the original demonstration from time step $t^*$ on $\tau^{\mathrm{demo}}_{t^* \rightarrow T}$. The fully augmented trajectory is then defined as $\tau_{i,k} = \tau^{\mathrm{aug}}_{i,k} \;\Vert\; \tau^{\mathrm{demo}}_{t^* \rightarrow T}$, where $\Vert$ denotes temporal concatenation. The final training dataset was obtained by aggregating all augmented trajectories across all spheres:
\begin{align}
\mathcal{D}_{\mathrm{aug}} = \bigcup_{i=1}^{I} \;\bigcup_{k=1}^{K} \;\tau_{i,k}.
\end{align}
The dataset $\mathcal{D}_{\mathrm{aug}}$ is used to train the policy, as described in \sref{sec:preliminaries}.
A summary of SART is presented in \aref{alg:sart}.

% algorithm
\begin{algorithm}[t]
\caption{Self-augmented robot trajectory (SART)}
\label{alg:sart}
\begin{algorithmic}[1]
\renewcommand{\algorithmicrequire}{\textbf{Input:}}
\renewcommand{\algorithmicensure}{\textbf{Output:}}
\REQUIRE A single demonstration $\mathcal{D}$ and annotated boundaries $\{(\kappa_i, r_i)\}_{i=1}^{I}$
\ENSURE Augmented training dataset $\mathcal{D}_{aug}$ and parameter of learned policy $\theta^{L}$
\STATE Initialize augmented dataset $\mathcal{D}_{\mathrm{aug}} \leftarrow \mathcal{D}$
\FOR{each key waypoint $i = 1$ to $I$}
    \FOR{$k = 1$ to $K$}
        \STATE Sample pose $\kappa_{i}^{k}$ on sphere with \eref{eq:rand_pos} and \eref{eq:rand_ori}
        \STATE Find convergence waypoint $\kappa^*_i \in \mathcal{W}_{t^*}$ with \eref{eq:target_pose}
        \STATE Record augment trajectory $\tau^{\mathrm{aug}}_{i,k}$: \\
        Moving the robot from $\kappa^*_i$ to $\kappa_{i}^{k}$ over time $\delta$ and then reversing the recorded sequence
        \STATE Aggregate dataset: $\mathcal{D}_{\mathrm{aug}} \leftarrow \mathcal{D}_{\mathrm{aug}} \cup \tau_{i,k}$, where $\tau_{i,k} = \tau^{\mathrm{aug}}_{i,k} \Vert \tau^{\mathrm{demo}}_{t^* \rightarrow T}$
    \ENDFOR
\ENDFOR
\STATE Learn $\theta^{L}$ by \eref{eq:bc_objective} on $\mathcal{D}_{\mathrm{aug}}$
\end{algorithmic}
\end{algorithm}

\section{Evaluation}
\label{sec:evaluation}

The SART framework was evaluated in both simulated and real environments with a focus on clearance-limited robotic tasks. 
The following three primary questions guided the analysis:
\begin{enumerate*}
    \item How does SART perform compared to baseline imitation learning methods when trained on clearance-limited robotic manipulation tasks?
    \item Which design of SART makes the best performance?
    \item How effective is SART in reducing overall human effort?
\end{enumerate*}

\subsection{Evaluation Setting}

\subsubsection{Environments and Tasks} \label{sec:eval:tasks}

\begin{figure}[t]
    \centering
    \begin{minipage}[t]{0.655\linewidth}
        \centering
        \textbf{Simulation Envs}\vspace{1mm}\\
        \begin{minipage}[t]{0.49\linewidth}
            \centering
            \includegraphics[width=\linewidth]{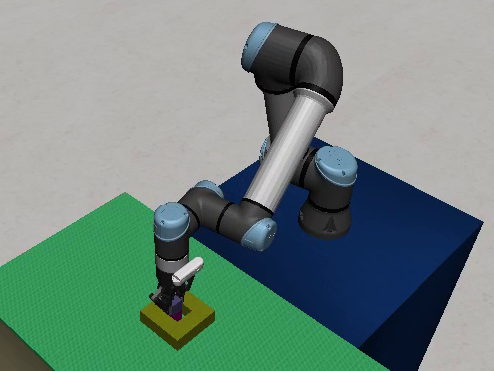}\\
            \footnotesize Peg-in-hole
        \end{minipage}
        \begin{minipage}[t]{0.49\linewidth}
            \centering
            \includegraphics[width=\linewidth]{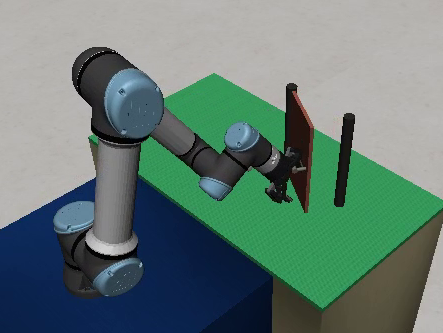}\\
            \footnotesize Door opening
        \end{minipage}
    \end{minipage}
    \begin{minipage}[t]{0.32\linewidth}
        \centering
        \textbf{Real-world Envs}\vspace{1mm}\\
        \includegraphics[width=\linewidth]{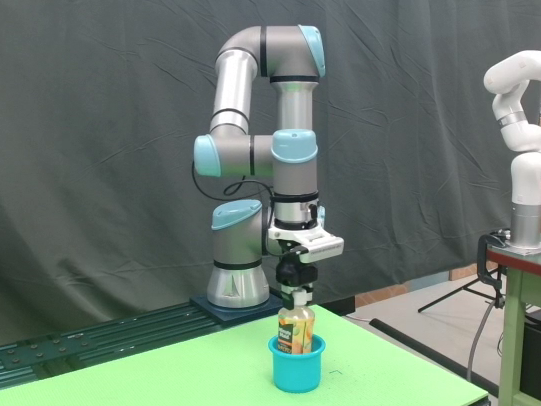}\\
        \footnotesize  Bottle placing
    \end{minipage}
    \\
    \vspace{1mm}
    \begin{minipage}[t]{0.32\linewidth}
        \centering
        \includegraphics[width=\linewidth]{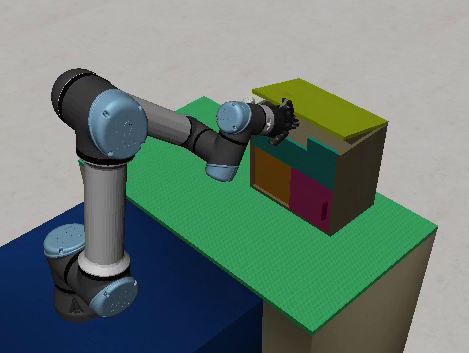}\\
        \footnotesize Lid opening
    \end{minipage}
    \begin{minipage}[t]{0.32\linewidth}
        \centering
        \includegraphics[width=\linewidth]{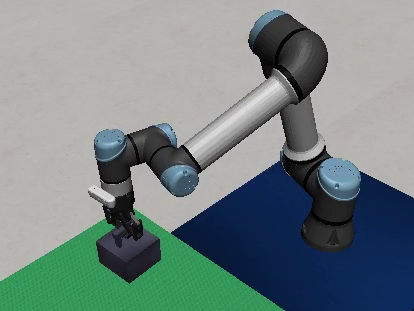}\\
        \footnotesize Toolbox picking
    \end{minipage}
    \begin{minipage}[t]{0.32\linewidth}
        \centering
        \includegraphics[width=\linewidth]{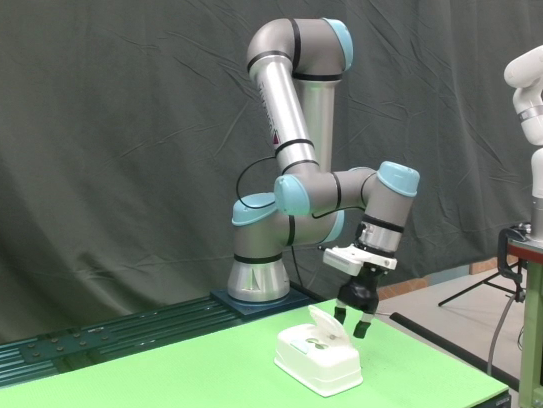}\\
        \footnotesize  Lid closing
    \end{minipage}
    \caption{Evaluation tasks. The four panels on the left show the simulated environments, where a robot performs the peg-in-hole, door opening, lid opening, and toolbox picking tasks. The two panels on the right show real-world environments, where a robot performs the bottle placement and lid closure tasks. These tasks represent diverse clearance-limited manipulation scenarios of varying complexity.}
    \label{fig:eval:tasks}
\end{figure}

Evaluation of the SART framework, 
as indicated in \fref{fig:eval:tasks}, 
was conducted in both simulated and real-world environments. The environment and task definitions are as follows:  

The \textbf{simulation environment} is built on the MuJoCo simulator \cite{todorov2012mujoco}, where the Universal Robots UR5e 6-DoF manipulator is equipped with a Robotiq 2F-85 gripper. 
Following four clearance-limited manipulation tasks were examined, each inherently involving collisions or tight physical interactions of varying complexity:
\begin{itemize}
    \item \textbf{Peg-in-hole:} An insertion task in which the robot inserts a $3\,\mathrm{cm}$ square peg into a $5\,\mathrm{cm}$ square hole, requiring precise alignment and repeated contact with the hole surface.  
    \item \textbf{Door opening:} A hinge-manipulation task in which the robot must hook and rotate a handle to open the door by at least $45^{\circ}$, inducing continuous interaction along the handle and hinge.  
    \item \textbf{Lid opening:} A task in which the robot lifts a hinged container lid to an angle of more than $120^{\circ}$, involving sustained interaction along the hinge joint.  
    \item \textbf{Toolbox picking:} A lifting task in which the robot must grasp the narrow handle of a toolbox placed on a table and raise it securely, requiring a precise approach and stable grasp.  
\end{itemize}

A \textbf{real-world environment} is set up with a UR5e robot equipped with a Robotiq Hand-E gripper. 
The following two daily manipulation tasks were performed, both requiring clearance-limited motion:
\begin{itemize}
    \item \textbf{Bottle placing:} A tight placing task in which the robot inserts a plastic bottle (diameter $6.5\,\mathrm{cm}$) into a cup (diameter $8.5\,\mathrm{cm}$, depth $5.5\,\mathrm{cm}$), often leading to interaction with the rim and inner surface of the cup.  
    \item \textbf{Lid closing:} A container-closing task in which the robot presses and fully closes the lid of a small plastic container, requiring accurate positioning and force application.  
\end{itemize}

The experiments employed a complete pipeline consisting of teleoperation, augmented data collection, policy training, and rollout implemented using a software framework called RoboManipBaselines \cite{murooka2025rmb}.
Teleoperation is performed using a 3D mouse, which commands relative end-effector translation/rotation and gripper open--close actions, at \SI{30}{\hertz}. The robot observations and actions were recorded at \SI{10}{\hertz} along with synchronized RGB images captured from a camera mounted on the gripper to stabilize policy learning.
During the policy rollout test, the goal objects of each task were randomly placed to validate generalizability. For the peg-in-hole method, the initial position of the hole was uniformly randomized within a $\pm 5\,\mathrm{cm}$ square region on the tabletop. For the remaining tasks, the initial positions of the objects (\eg door, lid, toolbox, and bottle) were randomized within a $\pm 10\,\mathrm{cm}$ square on the tabletop.

\subsubsection{Comparison Methods}

SART was evaluated on both simulated and real-world robotic manipulation tasks in comparison with the following baseline methods:

\begin{itemize}
    \item \textbf{Single-demo replay:} Direct replay of a single teleoperated trajectory without any additional learning or augmentation.  

    \item \textbf{Behavioral cloning (BC) \cite{bojarski2016end}:} Standard IL from multiple human demonstrations, referred to as BC, as described in \sref{sec:preliminaries}. The demonstrations were collected under randomized goal-object positions in accordance with each task setting.

    \item \textbf{Contact-free making IL easy with self-supervision (MILES) \cite{papagiannis2025miles}:} A self-augmented IL method from a single demonstration, based on MILES. Although MILES is closely related to SART, it differs in two critical ways—(i) the augmentation boundary is fixed, regardless of task geometry, with a sphere of $2\,\mathrm{cm}$ radius placed at every demonstration waypoint rather than selectively; and (ii) augmented trajectories are generated by returning to the center of each sphere, instead of progressing toward the convergence waypoint. In addition, unlike the original MILES formulation, which assumes access to costly contact-sensing systems for handling contact-rich scenarios, the comparison here is conducted under a lower-cost setup without contact sensing. Video results available at \url{https://sites.google.com/view/sart-il}.
\end{itemize}
Throughout this evaluation, $N$ denotes the total amount of training data (trajectories) used for policy learning. 
For BC, $N$ corresponds to the number of human demonstrations, whereas it includes both the initial human demonstration and robot-augmented trajectories for SART and contact-free MILES. 
All methods were trained with the same policy architecture and optimization setup (see Appendix~\sref{app:hypa}).

\subsection{Results}

\subsubsection{How does SART compare with baseline IL methods when trained on clearance-limited robotic manipulation tasks?} \label{sec:eval:comp_base}
To address this question, SART was evaluated against baseline IL methods on clearance-limited robotic manipulation tasks, and the results were analyzed both qualitatively and quantitatively.

\begin{figure}[t]
    \centering
    \includegraphics[width=\linewidth]{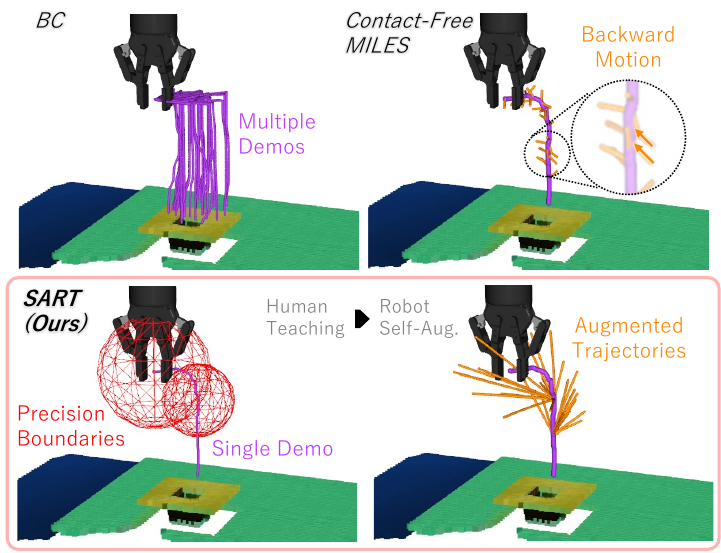}
    \caption{Comparison of training data ($N=30$) for the peg-in-hole task across varying methods. 
    \textbf{BC}: requires multiple human demonstrations (purple). 
    For clarity, the goal object (hole) is shown in a fixed position, although the demonstrations were collected with varying goal positions. 
    \textbf{Contact-free MILES}: augments trajectories from a single demonstration but produces backward motions with limited coverage (orange). 
    \textbf{SART (proposed)}: begins with a single human demonstration (purple) and annotated precision boundaries (red spheres) and subsequently  generates diverse, collision-free augmented trajectories (orange) autonomously.
    For clarity of the end-effector trajectories, the peg is omitted from the visualization.
    }
    \label{fig:eval:quali:demo}
\end{figure}

\begin{figure}[t]
    \centering
    \includegraphics[width=\linewidth]{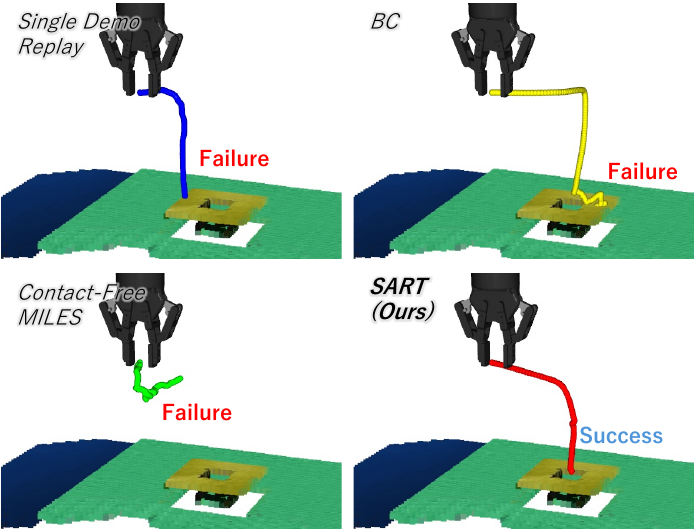}
   \caption{
    Rollouts of policies trained using $N=30$ training data across various methods. 
    Red: SART (proposed), Blue: Single-demo replay, Yellow: BC, Green: Contact-free MILES. 
    Among these trajectories, only SART completes the peg-in-hole task.
    For clarity of the end-effector trajectories, the peg is omitted from the visualization.
    }
    \label{fig:eval:quali:rollout}
\end{figure}

\textbf{Qualitative analysis} 
A visual inspection of the demonstrations and rollouts revealed fundamental differences across the baselines, as shown in \fref{fig:eval:quali:demo} and \fref{fig:eval:quali:rollout}.
Single-demo replay consistently fails under randomized goal-object positions because it lacks adaptation beyond the demonstrated trajectory. BC collects multiple human demonstrations to generalize new object placements; however, the learned policy suffers from compounding errors during execution and failures before task completion. Contact-free MILES struggles because the augmentation boundaries are uniformly small and include backward motion. Consequently, policies learned with contact-free MILES often stall mid task and fail to achieve their goals. In contrast, SART generates diverse goal-directed augmented trajectories. By adaptively expanding the precision boundaries only around critical waypoints, SART avoids backward motion and encourages progress toward the object, resulting in consistent task completion.

\textbf{Quantitative analysis} 
The success rates obtained in both simulated and real-world environments are summarized in \tref{tab:eval:quanti}. 
Single-demo replay yields the lowest performance owing to its lack of robustness to object position variations. 
Contact-free MILES performs even worse as backward motions are inadvertently included in the training trajectories, resulting in a poor success rate for the overall tasks. 
BC achieves higher success than single-demo replay and demonstrates some generalization to randomized goal-object placement; however, its average performance remains below $50\%$ because of compounding errors, which is consistent with qualitative observations. 
SART achieved the highest success rates across all environments, with statistically significant improvements across all baselines. In both simulation and real-world experiments, SART delivered nearly double the performance of BC, establishing it as the most reliable approach for clearance-limited manipulation tasks.

\begin{table*}[t!]
\caption{
Quantitative comparison of SART against baselines on clearance-limited manipulation tasks. 
In simulation, each method’s success rate is computed per task from $100$ test executions: 
$10$ policies are trained with different random seeds ($N=40$ training trajectories per policy), 
and evaluated using $10$ rollouts. 
In the real-world setup, results are averaged per task over $30$ executions--- 
$3$ independently trained policies ($N=30$) $\times$ $10$ rollouts. 
Single-demo replay does not involve policy training; its success rate is obtained by replaying 
the single demonstration $100$ times in simulation and $10$ times in the real-world environment, per task. \textbf{Bold} indicates the best performance. 
``Avg.'' denotes the mean $\pm$ standard deviation across tasks; an asterisk ($^*$) indicates results significantly lower than those of SART ($t$-test, $p < 0.05$).
}
\label{tab:eval:quanti}
\centering
\vspace{1mm}
\resizebox{\textwidth}{!}{% <------ Don't forget this %
\begin{tabular}{p{3.5 cm} C{1.9 cm} C{2.3 cm} C{2 cm} C{2.7 cm} C{0.01 cm} C{2.4 cm} C{1.9 cm} C{1.8 cm}}
\hline
Methods & \multicolumn{4}{c}{Simulation task success rate [\%]} & & \multicolumn{2}{c}{Real task success rate [\%]} & Avg. [\%] \\ 
\cline{2-5} \cline{7-8}
& Peg-in-hole & Door opening & Lid opening & Toolbox picking & & Bottle placing & Lid closing &   \\ \hline
Single-demo replay   & $4$   & $6$   & $28$  & $16$  & & $0$   & $30$  &  $14^{*} \pm 11$ \\
BC                   & $28$  & $58$  & $31$  & $8$   & & $67$ & $60$  &  $36^{*} \pm 26$ \\
Contact-free MILES   & $41$  & $0$   & $0$   & $2$   & & $53$  & $43$  &  $16^{*} \pm 24$ \\
\textbf{SART (Ours)}          & $\mathbf{89}$ & $\mathbf{88}$ & $\mathbf{83}$ & $\mathbf{67}$ & & $\mathbf{80}$ & $\mathbf{93}$  & $82 \pm 18$ \\ \hline
\end{tabular}
}
\end{table*}

\begin{figure}[t]
    \centering
    \includegraphics[width=\linewidth]{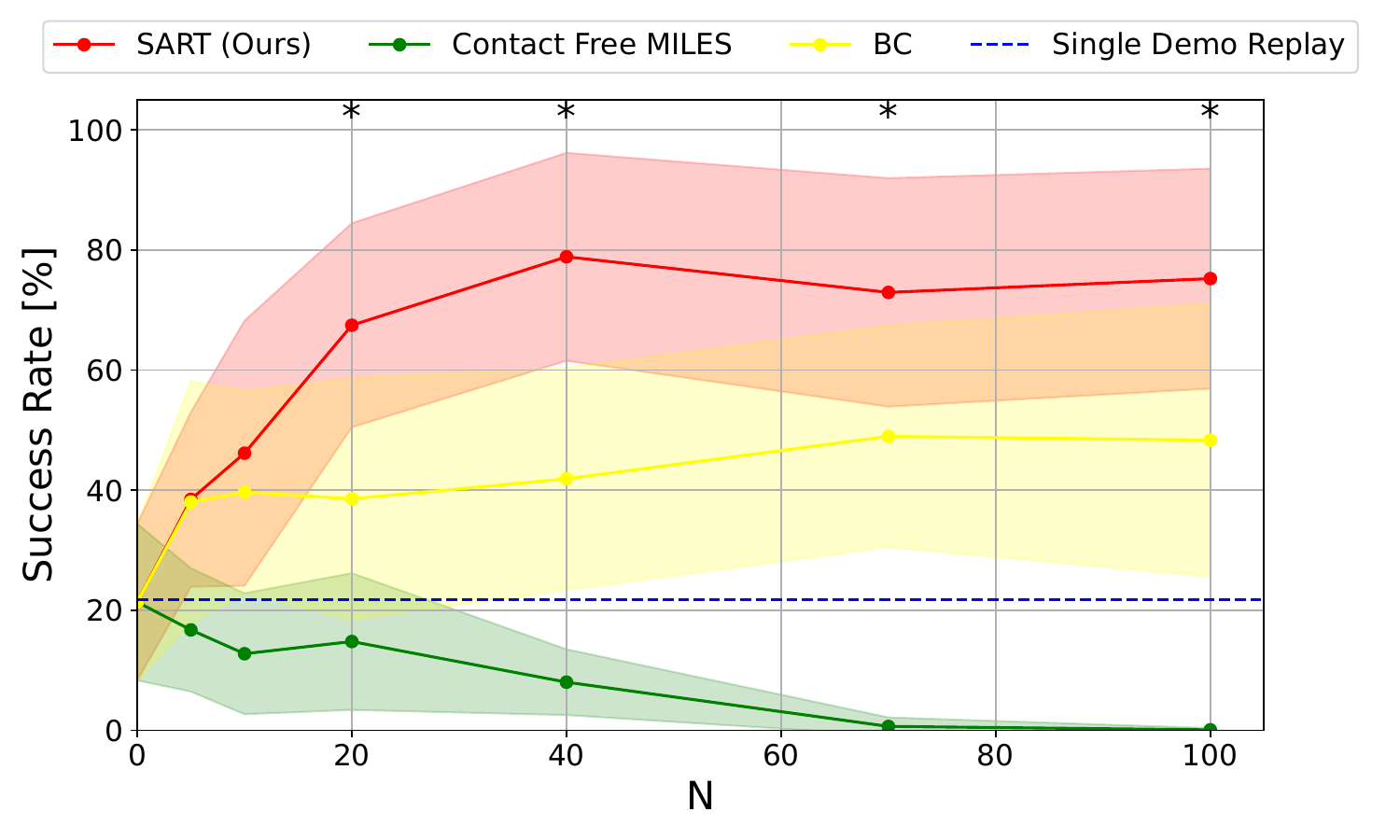}
    \caption{
    Comparison of success rates with respect to the number of training data ($N$) for SART and baseline methods on clearance-limited robotic manipulation tasks. 
    Only simulation tasks are evaluated. 
    Each result is averaged over $10$ independently trained policies (with different random seeds), where each policy is tested with $10$ rollouts per task, giving $100$ executions per method. 
    For single-demo replay, no policy is trained; its success rate is obtained by replaying a single demonstration $100$ times during evaluation.
    An asterisk ($^*$) indicates results where SART significantly outperforms all baselines ($t$-test, $p < 0.05$).
    }
    \label{fig:eval:quanti}
\end{figure}
The success rate trend across varying dataset sizes is shown in \fref{fig:eval:quanti}. 
Contact-free MILES degrades rapidly, approaching near-zero success as the number of augmented trajectories increases, because of the growing presence of backward motions in the training data. BC benefits from additional demonstrations but plateaus below a $60\%$ success rate. 
In contrast, SART scales more effectively: when $N \geq 20$ training data are available, the performance gap between SART and the baselines becomes statistically significant and remains consistently superior as the dataset size increases. Importantly, this strong performance suggests that SART’s augmented trajectories carry values comparable to those of additional human demonstrations, highlighting its effectiveness in reducing human effort.

\subsubsection{Which design of SART makes the best performance?}
\begin{figure}[t]
    \centering
    \includegraphics[width=\linewidth]{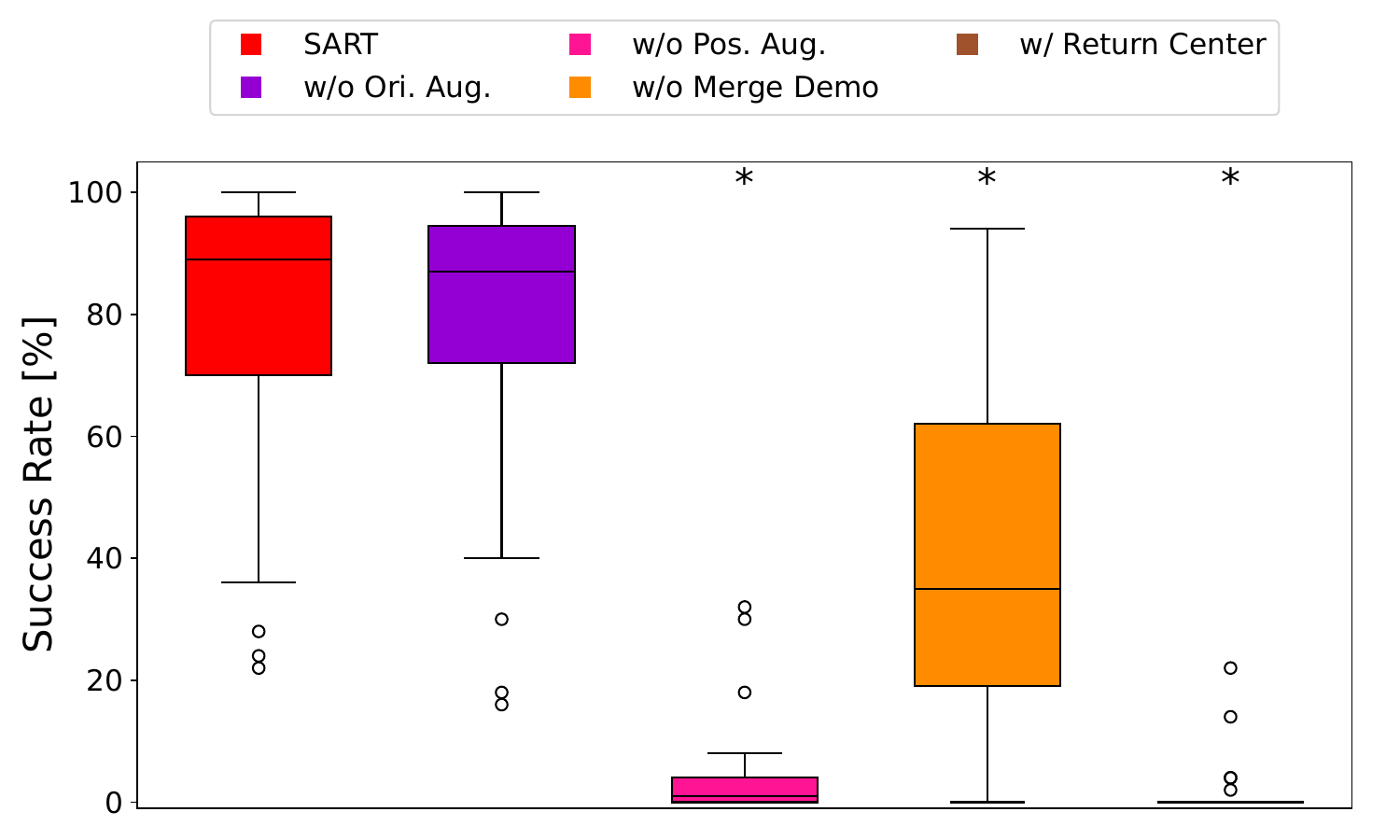}
    \caption{
    Comparison of SART with ablated variants. Only simulation results are reported. Each result is averaged over $10$ independently trained policies (with different random seeds) and $N = 100$ training trajectories, where each policy is tested with $10$ rollouts, yielding $100$ executions per method. An asterisk ($^*$) indicates cases where removing a SART design component leads to a statistically significant performance drop ($t$-test, $p < 0.05$).
    }
    \label{fig:eval:ablation}
\end{figure}
To better understand which design elements are most critical to SART’s performance, an ablation study was conducted by systematically removing or modifying the individual components. As shown in \fref{fig:eval:ablation}, three ablations result in significant performance degradation, whereas one yields only a minor drop.

\textbf{Without position augmentation (w/o Pos. Aug.).} 
When position augmentation is disabled (\ie $\mathbf{p}_{i}^{k} = \mathbf{c}^{\mathrm{pos}}_i$), the data coverage becomes similar to that of BC, providing limited variability. Consequently, the trained policy suffers from compounding errors and fails in several tasks, showing a statistically significant decrease compared with the full SART.

\textbf{With return to center (w/ Return center).} 
In this variant, the augmentation trajectory is connected to the sphere center rather than the convergence waypoint (\ie $\kappa_{i}^{*} = \kappa_{i}$). This variant often induces backward motions similar to those observed in contact-free MILES, which cause policies to stall mid-tasks, resulting in a significant performance drop.

\textbf{Without merging demonstration (w/o merge demo).} 
Excluding the suffix of the original demonstration (\ie $\tau_{i,k} = \tau^{\mathrm{aug}}_{i,k}$) after each augmentation trajectory biases the collected data toward the augmented regions. Consequently, the model was insufficiently trained in the later stages of the task and failed to complete rollouts reliably, resulting in a significant performance drop.

\textbf{Without orientation augmentation (w/o Ori. Aug.).} 
In this setting, the orientation is fixed (\ie $\mathbf{R}_{i}^{k} = \mathbf{C}^{\mathrm{ori}}_i$), and only positional variation is introduced. Because the evaluation tasks primarily assessed positional generalization, this ablation resulted in only a small degradation. Nevertheless, the full SART still outperformed this variant, confirming the benefit of including randomization in both the position and orientation.

These ablation results demonstrated position augmentation, merging with the original demonstration, and a progression-based augmentation design (rather than returning to sphere centers) are essential for SART’s performance gains. 
Although orientation augmentation is less critical in the current benchmarks, it provides robustness and supports the overall generality of the framework.
Further experiments were conducted to further analyze alternative convergence strategies and sampling schemes (\apref{app:add:ablation}), as well as to compare SART with contact-free MILES variants using different fixed sphere sizes (\apref{app:add:miles}).

\subsubsection{How effective is SART in reducing overall human effort?}
\begin{figure}[t]
    \centering
    \includegraphics[width=\linewidth]{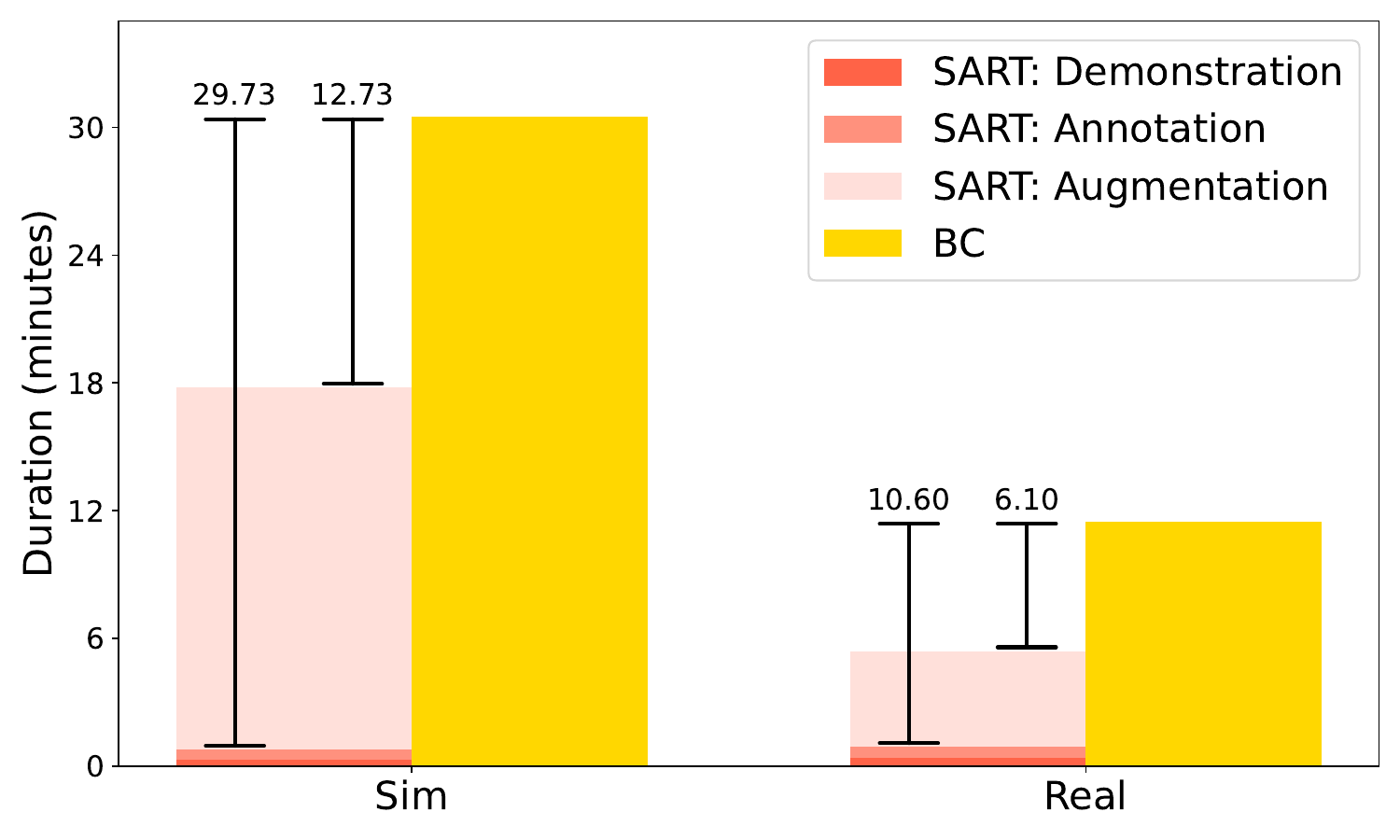}
    \caption{
    Comparison of average duration for dataset collection in both simulation and real-world settings. Results are averaged across all simulation tasks (for $N=100$ trajectories) and real-world tasks (for $N=30$ trajectories). For BC, the duration corresponds to collecting $N$ full demonstrations. For SART, the total duration consists of three parts: "demonstration", the average time for collecting a single demonstration; "annotation", the average time over three annotation trials; and "augmentation", the time required to generate $N$ augmented trajectories.
    }
    \label{fig:eval:duration}
\end{figure}
The evaluation quantifies the overall human effort via dataset-collection duration. As shown in \fref{fig:eval:duration}, SART requires substantially less human involvement than BC. In simulation---where no environmental reset is needed---the average duration was $17.77\,\mathrm{min}$ for SART versus $30.50\,\mathrm{min}$ for BC. In real-world tasks---which do require manual resetting---the average duration is $5.40\,\mathrm{min}$ for SART and $11.50\,\mathrm{min}$ for BC.

A component-wise breakdown clarifies the savings. When considering only manual processes (demonstration and annotation), SART reduced human effort relative to BC by $29.73\,\mathrm{min}$ in the simulation and $10.60\,\mathrm{min}$ in real-world settings. Even after including robot self-augmentation, the total duration remained low, with reductions of $12.73\,\mathrm{min}$ (simulation) and $6.10\,\mathrm{min}$ (real).

These results indicate that SART lowers the human effort required for data collection and shortens the total duration needed to construct a dataset, while maintaining significantly higher success rates, as shown in \sref{sec:eval:comp_base}.

\section{Discussion}
\label{sec:discussion}

SART combines a single-human demonstration with lightweight precision-boundary annotation and subsequent \emph{robot self-augmentation}. Across clearance-limited manipulation tasks in simulation and the real world, SART yielded higher success rates than baseline IL methods and reduced the human duration required to generate training datasets (see \sref{sec:evaluation}). Together, these results suggest that precision-aware augmentation can expand demonstration coverage without compromising safety or increasing the human workload.

Although SART demonstrated significantly better performance than behavioral cloning (BC) even when trained from a single demonstration, this result may partly stem from the use of a single trajectory that avoids the inherent ambiguity of human demonstrations. In IL, human demonstrations often represent multi-valued mappings, where multiple distinct trajectories can achieve the same goal. In this work, SART focuses on validating the efficacy of single-demonstration augmentation using a simple multilayer perceptron (MLP)-based policy; however, such a model is limited in its ability to represent multiple optimal strategies demonstrated by humans. Extending SART to augment and learn from a small set of diverse human demonstrations could further amplify its performance while addressing this multimodal challenge. To achieve this, future work should explore more expressive policy architectures, such as diffusion-based policies~\cite{chi2023diffusion}, which are capable of capturing multiple feasible action distributions and thus better leverage diverse human data.

Another limitation is the manual effort required for annotation. Although the overall effort has been substantially reduced, the boundary specification still relies on humans. Possible extensions include auto-suggesting precision boundaries with collision risk assessment using 3D perception and robot kinematics \cite{zucker2013chomp} or using human behavioral characteristics \cite{Oh2024DPIIL}. Such approaches can further reduce the annotation overhead while preserving safety guarantees.

An object-centric hand–eye view also provides robustness to global scene variations because tasks can be solved as long as the goal object remains visible. However, this introduces a visibility constraint; if the goal object is not observed by the wrist-mounted camera, the policy cannot solve the task. To overcome this limitation, future work may explore multi-view augmentation setups that combine hand–eye and third-person perspectives \cite{mandlekar2023mimicgen} to diversify the training distribution.

Beyond these task-specific extensions, a broader opportunity lies in scaling SART toward large-scale robot learning. Recent advances in robotic foundational models~\cite{o2024rtx, kim2025openvla} have shown that large, cross-task datasets leveraging broad pre-trained visuomotor representations can substantially reduce the number of demonstrations required for new tasks. While these models rely on massive and diverse datasets to achieve generalization, the proposed SART framework can be viewed as a complementary approach that supports scalable and safe data collection at the per-task level. By autonomously transforming a small number of human demonstrations into diverse, collision-free trajectories, SART has the potential to contribute to the development of scalable datasets beneficial for such models. As a potential extension, SART could be expanded to increase data diversity in simulation—\eg by varying object sizes~\cite{lin2025cpgen} or poses~\cite{yu2025r2r2r}—which may help support broader generalization in robot learning.

While SART enables safe data augmentation within annotated precision boundaries, the manual effort required to specify these regions may still impose an additional burden on the human operator. To further reduce this overhead, a promising extension would be to incorporate classical motion-planning approaches that automatically generate collision-free trajectories in contact-free regions~\cite{valassakis2021coarse, Sundaralingam2023curobo}. Such integration could complement SART’s safety guarantees by automating the exploration process without relying on detailed human annotations.

Despite the safety achieved in such contact-free augmentation, SART does not explicitly handle contact, and its applicability to contact-rich tasks remains limited. A future direction is to develop a unified framework that integrates the proposed contact-free SART with contact-aware augmentation methods~\cite{papagiannis2025miles}, enabling safer, more reactive data generation in contact-rich scenarios.

\section{Conclusions}
\label{sec:conclusions}

This study introduced SART, a framework for IL that requires only a single human demonstration with a lightweight annotation of precision boundaries. By combining one-time human teaching with robot self-augmentation, SART generates diverse collision-free trajectories that improve both the safety and efficiency of data collection.
Experiments across simulated and real-world clearance-limited tasks demonstrated that SART achieved significantly higher success rates than baselines, such as BC and contact-free MILES, while requiring less human effort. 

\section*{Acknowledgement}

This research was conducted using the financial support and experimental facilities provided by the National Institute of Advanced Industrial Science and Technology (AIST). We acknowledge the support and assistance provided throughout this study.

% \section*{Funding} ?

%%%%%%%%%%%%%%%%%%%%%%%%%%%%%%%%%%%%%%%%%%%%%%%%%%%%%%%%%%%%%%%%%%%%%%%%%%%%%%%%
\bibliographystyle{tfnlm}
\bibliography{ref}

%%%%%%%%%%%%%%%%%%%%%%%%%%%%%%%%%%%%%%%%%%%%%%%%%%%%%%%%%%%%%%%%%%%%%%%%%%%%%%%%
\newpage
\appendix
\section{Hyperparameter Setting}
\label{app:hypa}
All comparison methods, including SART, were trained using a multilayer perceptron (MLP) policy head. The MLP architecture and the training hyperparameters are summarized in \tref{tab:app:param}. The visual observations were encoded using a ResNet18 backbone, which was fine-tuned on our dataset and concatenated with the gripper state features. The entire policy network is optimized using the behavioral cloning objective in \eref{eq:bc_objective}. One training epoch corresponds to an iteration over the entire training dataset.

\begin{table*}[t!]
\caption{Hyperparameter setting
}
\label{tab:app:param}
\centering
\vspace{1mm}
\resizebox{0.5\textwidth}{!}{% <------ Don't forget this %
\begin{tabular}{p{6cm} C{3cm}}
\hline
Hyperparameters & Value\\
\hline
Optimizer & AdamW\\
Activation function & ReLU\\
Number of hidden layers & 2\\
Number of hidden units per layer & 512\\
Dimension of state feature & 512\\
Dimension of image feature & 512\\
Observation history & 2\\
Minibatch size & 32\\
Learning rates & 1e-5\\
Weight decay & 1e-4\\
Training epoch & 40\\
Augmentation duration $\delta$ & 2.0 [s]\\
\hline
\end{tabular}
}
\end{table*}

\section{Annotation Interface Implementation}
\label{app:annotation}

An interactive interface was developed to facilitate the intuitive annotation of precision boundaries. 
During the demonstration, a 3D point cloud of the scene was reconstructed using a wrist-mounted RGB-D camera. 
The demonstration trajectory was then visualized using a virtual robot model, which showed the end-effector pose at each waypoint in 3D. 
For annotation, the system highlights each selected key waypoint in the sequence and renders a transparent sphere centered at the end-effector position. 
The annotator can adjust the radius $r_i$ interactively using simple keyboard inputs (increase/decrease) while viewing the overlap between the sphere and the surrounding point cloud to avoid collisions. 
This design enables the annotator to focus on one waypoint at a time, thereby reducing cognitive load. 
By stepping through the trajectory in order, the annotation process becomes straightforward; the user inspects the local geometry, adjusts the boundary radius, and proceeds to the next waypoint. 
In practice, this procedure allows annotation of an entire trajectory within a few minutes with minimal effort.

\begin{figure}[t]
    \centering
    \includegraphics[width=\linewidth]{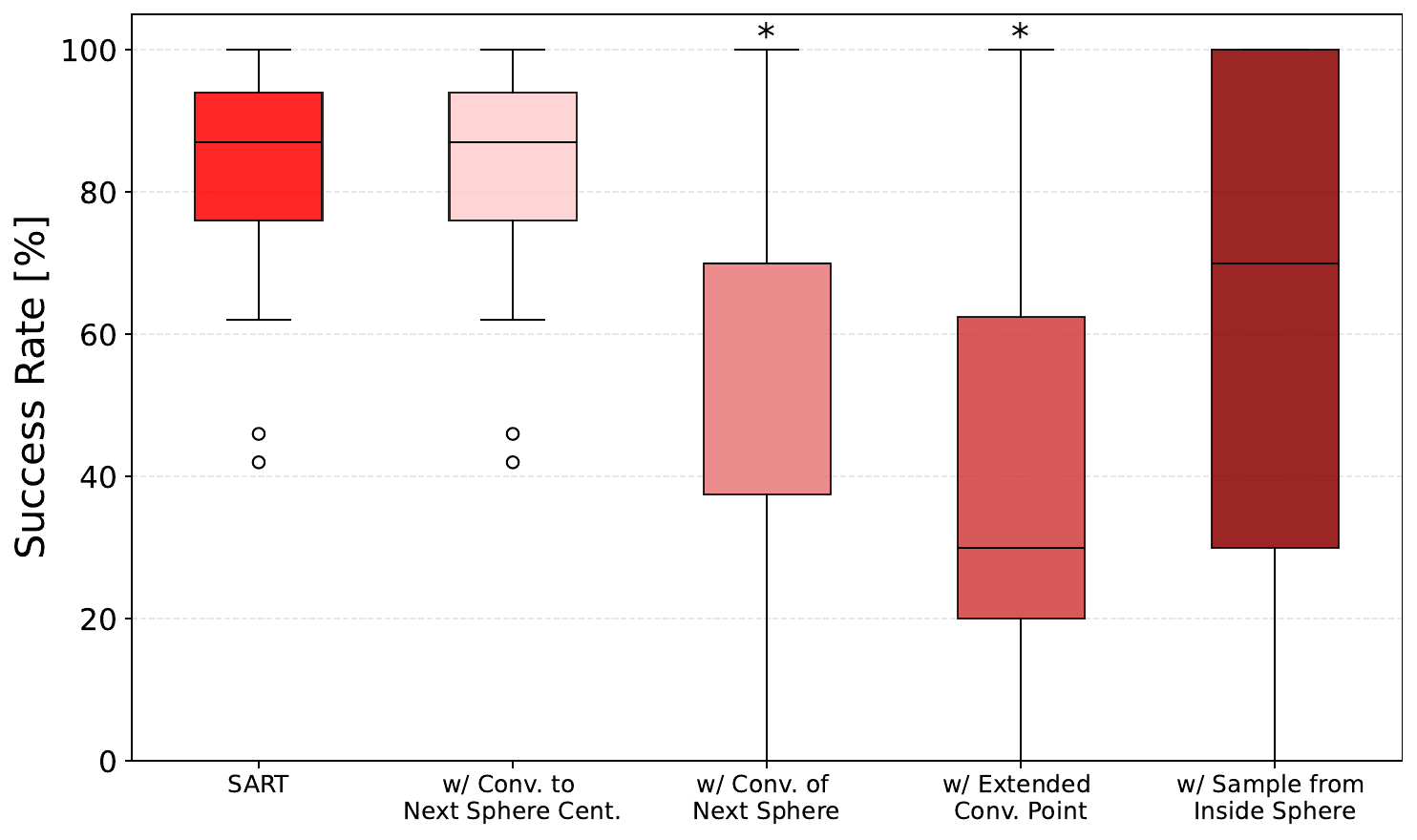}
    \caption{
    Comparison of SART with ablated variants. Only simulation results are reported. Each result is averaged over $5$ independently trained policies (with different random seeds) and $N = 40$ training trajectories, where each policy is tested with $10$ rollouts, yielding $50$ executions per method. An asterisk ($^*$) indicates cases where removing a SART design component leads to a statistically significant performance drop ($t$-test, $p < 0.05$).
    }
    \label{fig:eval:add:ablation}
\end{figure}

\section{Additional Results: SART Ablations}
\label{app:add:ablation}
To further analyze the design factors that contribute to SART’s performance, additional ablation experiments were conducted to evaluate  
\begin{enumerate*}
    \item the effect of alternative convergence strategies and  
    \item the impact of sampling positions inside the precision sphere rather than on its surface.  
\end{enumerate*}
These analyses clarify how SART’s convergence and sampling formulations affect safety and efficiency during data augmentation.

As shown in \fref{fig:eval:add:ablation}, several variants were tested:
\begin{itemize}
    \item \textbf{Convergence to Next-Sphere Center (w/ Conv. to Next Sphere Cent.)}  
    The robot converges to the center of the next precision sphere $\boldsymbol{c}^{\mathrm{pos}}_{i+1}$ instead of the current convergence point $\boldsymbol{c}^{\mathrm{pos}}_{t^*}$.  
    Because the annotated spheres are always positioned consecutively under the current annotator, this variant yields performance identical to the default SART.  
    Note that this arrangement was not intentionally constrained.
    \item \textbf{Convergence to Next-Sphere’s Convergence Point (w/ Conv. of Next Sphere)}  
    The convergence waypoint is changed to the next sphere’s convergence point $\boldsymbol{\kappa}^{*}_{i+1}$ instead of the current sphere’s $\boldsymbol{\kappa}^{*}_{i}$.  
    This modification largely degrades the smoothness of the trajectory, leading to unstable transitions and reduced policy performance.
    \item \textbf{Extended Convergence Point (w/ Extended Conv. Point)}  
    In this variant, the convergence point is set twice as far from the sphere center, i.e.,  
    $\left\| \mathbf{c}^{\mathrm{pos}}_{t} - \mathbf{c}^{\mathrm{pos}}_{i} \right\|_2^2 > (2r_i)^2$ in Eq.~\eqref{eq:target_pose}.  
    This extension frequently induces collisions during self-augmentation, which leads to the inclusion of multiple failed trajectories in the training dataset and consequently causes a significant performance degradation.
    \item \textbf{Sampling from Inside the Sphere (w/ Sample Inside Sphere)}  
    Random poses $\boldsymbol{p}^{k}_{i}$ are sampled inside the precision sphere rather than on its surface,  
    $\left\| \boldsymbol{p}^{k}_{i} - \boldsymbol{c}^{\mathrm{pos}}_{i} \right\|_2 < r_i$.  
    Because this reduces the spatial range of exploration, the resulting dataset lacks diversity, producing poorer generalization compared with the default configuration.
\end{itemize}
Overall, these results confirm that SART’s design—particularly the use of adaptive precision boundaries and convergence to the nearest feasible waypoint—plays a crucial role in balancing safety and efficiency during augmentation.

\begin{figure}[t]
    \centering
    \includegraphics[width=\linewidth]{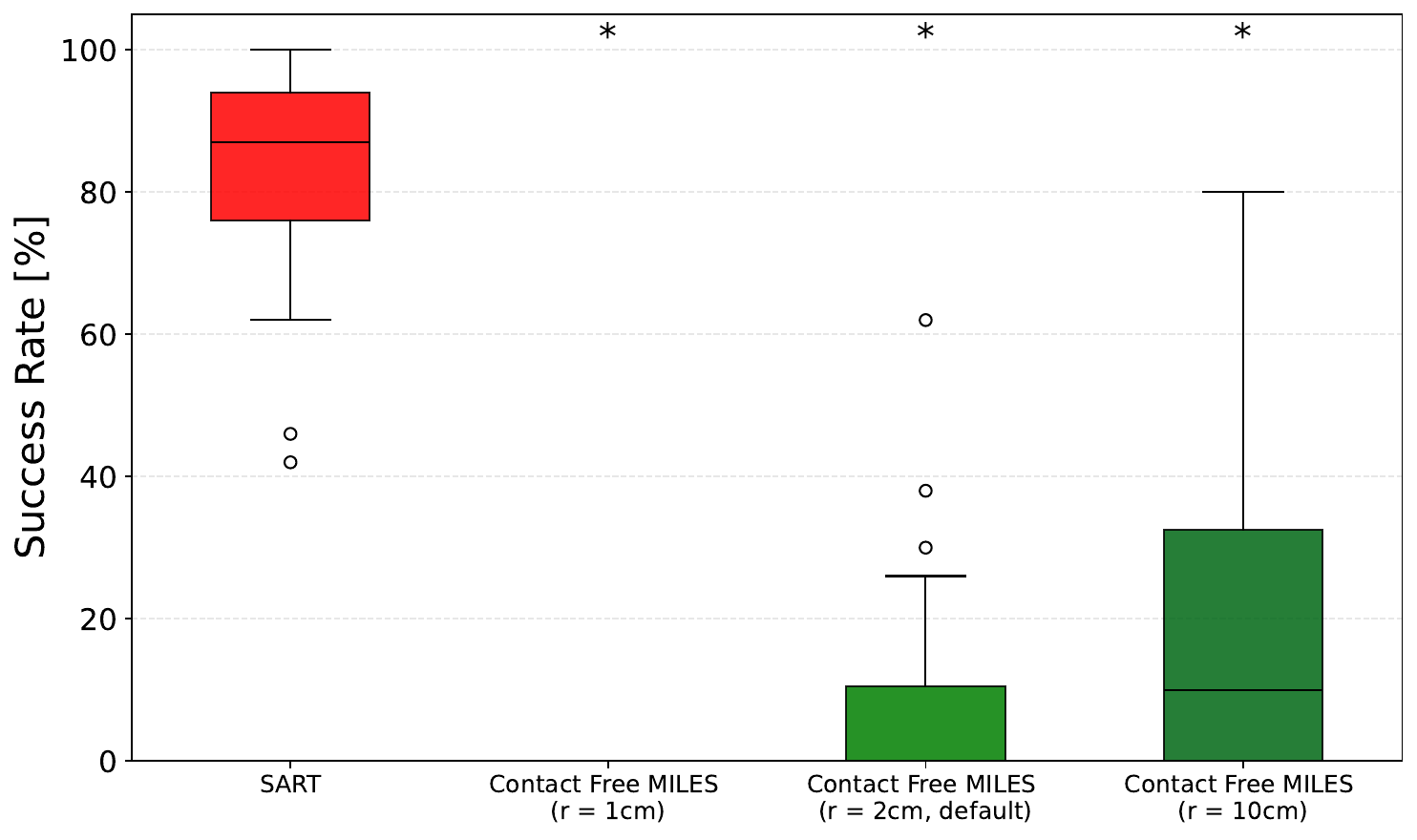}
    \caption{
    Comparison of SART with contact-free MILES variants using different fixed sphere sizes. Simulation results are reported.
    Each result is averaged over $5$ independently trained policies (with different random seeds) and $N=40$ training trajectories, where each policy is tested with $10$ rollouts, yielding $50$ executions per method.
    An asterisk ($^*$) indicates cases where a configuration yields statistically lower performance than SART ($t$-test, $p<0.05$).
    }
    \label{fig:eval:add:miles_variants}
\end{figure}

\section{Additional Results: Comparison between SART and Contact-Free MILES Variants}
\label{app:add:miles}
To further analyze the effect of the augmentation boundary size, additional experiments were conducted comparing SART with contact-free MILES variants that used different fixed sphere radii: $1\,\mathrm{cm}$, $2\,\mathrm{cm}$ (default), and $10\,\mathrm{cm}$. These settings evaluate whether reducing or enlarging the augmentation region influences policy robustness in clearance-limited tasks.

As shown in \fref{fig:eval:add:miles_variants}, all contact-free MILES variants—regardless of sphere size—exhibited significantly poorer performance than SART. Increasing the fixed sphere radius in contact-free MILES from $1\,\mathrm{cm}$ to $10\,\mathrm{cm}$ led to a slight improvement in performance due to broader exploration, but the results remained substantially below those achieved by SART. When the sphere size was too small ($1\,\mathrm{cm}$), the robot explored an overly restricted region, resulting in limited data diversity and weaker generalization. Conversely, when the sphere size was excessively large ($10\,\mathrm{cm}$), most of the sampled poses fell outside the feasible collision-free region, leaving only a small portion of the augmentation space effectively usable. As a result, the generated dataset became sparse and unbalanced, leading to degraded task success. In contrast, SART adaptively defines precision boundaries according to task-specific spatial constraints, maintaining both collision safety and trajectory diversity. These results confirm that SART’s adaptive, boundary-aware formulation is substantially more effective and robust than globally fixed augmentation regions.

\end{document}